\documentclass[sigconf,screen]{acmart}

\acmConference[CAIN 2024]{3rd International Conference on AI Engineering — Software Engineering for AI}{April 2024}{Lisbon, Portugal}

\AtBeginDocument{%
  \providecommand\BibTeX{{%
    \normalfont B\kern-0.5em{\scshape i\kern-0.25em b}\kern-0.8em\TeX}}}

\graphicspath{{figures}}
\usepackage{hyperref}
\usepackage{blindtext}
\renewcommand\footnotetextcopyrightpermission[1]{}

\begin{document}

\title{Prompt Smells: An Omen for Undesirable Generative AI Outputs\\}

\author{Krishna Ronanki}
\email{krishna.ronanki@gu.se}
\orcid{0009-0001-8242-6771}
\affiliation{%
  \institution{University of Gothenburg}
  \streetaddress{Hörselgången 5}
  \city{Gothenburg}
  \country{Sweden}
  \postcode{417 56}
}

\author{Beatriz Cabrero-Daniel}
\email{beatriz.cabrero-daniel@gu.se}
\orcid{0000-0001-5275-8372}
\affiliation{%
  \institution{University of Gothenburg}
  \streetaddress{Hörselgången 5}
  \city{Gothenburg}
  \country{Sweden}
  \postcode{417 56}
}

\author{Christian Berger}
\email{christian.berger@gu.se}
\orcid{0000-0002-4828-1150}
\affiliation{%
  \institution{University of Gothenburg}
  \streetaddress{Hörselgången 5}
  \city{Gothenburg}
  \country{Sweden}
  \postcode{417 56}
}

\renewcommand{\shortauthors}{Ronanki et al.}


\maketitle

\section*{Generative Artificial Intelligence}

Recent trends in the world of Generative Artificial Intelligence (GenAI) focus on developing deep learning (DL)-based models capable of learning structures and temporal patterns from supplied training data to generate content in different formats like text, images, or sound. GenAI models have been widely used in various applications, including creating stories, illustrations, poems, articles, computer code, music compositions, and videos~\cite{10.1145/3422622, ruthotto2021introduction}.

A critical limitation of such GenAI models is a phenomenon called extrinsic hallucinations. Hallucinations are instances where GenAI systems generate content that is unrealistic or nonsensical, often inconsistent with the provided context or training data~\cite{bang2023multitask}. This phenomenon poses a threat to ongoing efforts to generalise the application of GenAI to non-entertainment contexts like software development or even high-risk AI-based solutions, where such AI/AI-enabled systems have the potential to adversely impact people's safety, health, or fundamental rights~\cite{hupont2023documenting}. This can lead to significant challenges in achieving and maintaining trustworthiness of GenAI~\cite{dwivedi2023so}, particularly in light of GenAI's increasing popularity and the growing inclination of individuals and organisations to explore and expand the boundary of its use cases. 

\section*{Desirability of Generative AI Outputs}

The concept of desirable outputs was briefly mentioned in existing literature like ~\cite{10.1145/3491102.3501825} but has not been properly defined or explored enough.

\vspace{0.2cm}
\noindent\fbox{%
\begin{minipage}{.96\columnwidth}
\textbf{We define desirability as a quality property of GenAI outputs that depends on three factors: 1) The accuracy or correctness of the information in the output, 2) The adherence to the ideal format for the output, and 3) The relevance of the output to the specific context of the task or instruction given through the prompt.}
\end{minipage}
}\\

\begin{figure} [ht!]
    \centering
    \includegraphics[width=1.0\columnwidth]{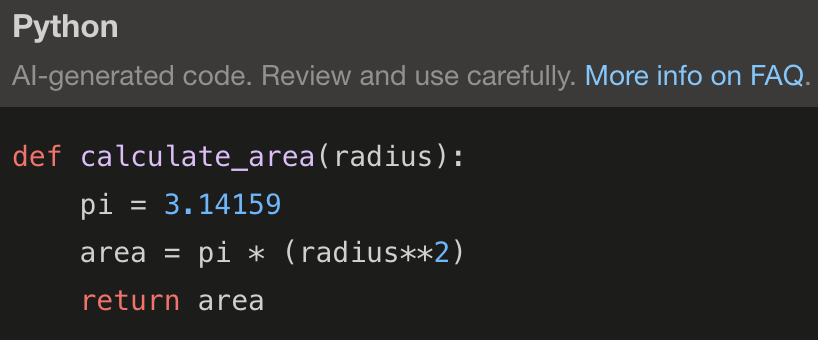}
    \caption{Desirable GenAI output}
    \label{fig:1}
\end{figure}

Consider the AI-generated code block presented in Figure \ref{fig:1}: \emph{Accuracy or correctness} refers to how well the output matches the reality or the truth. In the context of this Python code, the accuracy is high because it correctly calculates the area of a circle. The value of pi is also accurately represented up to five decimal places. \emph{Format Adherence} refers to how well the output follows the ideal format for the task or instruction. The Python code adheres to the correct syntax and structure of a Python function. It defines a function ``calculate\_area'' that takes a parameter ``radius'', calculates the area of a circle, and returns the result. The code is properly indented and easy to read, which is a key aspect of Python’s formatting guidelines. \emph{Relevance} refers to how well the output relates to the specific context of the task or instruction. If the task was to write a Python function to calculate the area of a circle given its radius, then this code is highly relevant as it performs exactly this task.

While we acknowledge that there is no single definition that can fully capture the complexity of desirability as a quality property, we believe that our definition is the most comprehensive one available to date. It covers aspects that are often overlooked or implicitly assumed, but never mentioned. We hope that our work will contribute to the ongoing conversation about the desirability of generative AI outputs and help advance the field in a meaningful way.

\section*{Prompt Engineering}

To generate such desirable outputs using a GenAI model, one typically provides natural language (NL) text such as questions or tasks as input within a \emph{prompt} for content generation. Such input can be a few words, a sentence, or any other form of text that is considered appropriate for the task. The AI model takes this input and generates possible outputs based on its previous experience and the patterns acquired during training. The emerging practice of utilising carefully selected and composed NL instructions to achieve a desirable output from a generative AI model is called \emph{prompt engineering}~\cite{10.1145/3491102.3501825}.

Prompt engineering plays a crucial role as the selection of prompts can have a significant impact on downstream tasks, in particular for the zero-shot setting~\cite{perez2021true}, where NL sentences are used only to describe the problem or the desired output without providing any examples~\cite{10.1145/3544549.3585737}. However, the problem with NL is that it can be ambiguous in some contexts and NL output generated by LLMs is non-deterministic. Even in few-shot strategies, where examples of the desired outputs are provided, the order, in which the samples are presented to the model, often makes a difference between near state-of-the-art and random guess performance~\cite{lu2021fantastically}.

Experts are advocating in favour of standardising NL prompt structures~\cite{gwern2023}, as the model’s output relevance to the expected result was observed to improve the more it is exposed to inputs of a similar nature for a specific task. However, to properly define and establish a structure that can be generalised, it is crucial to identify any underlying patterns within the prompts that consistently generate desirable output for a given task. Different prompts need to be systematically evaluated in order to assess what different outputs they may lead to and to reverse engineer how these prompts are \emph{understood} by LLMs, thereby discovering any discrepancies between what is assumed or expected and what is understood by the models~\cite{cheng2023prompt}. This was the provenance of prompt patterns. Prompt patterns are codified domain-independent reusable patterns that can be applied to the input prompts to improve the robustness of the user's interaction with a GenAI model~\cite{white2023prompt}.

The emergence of GenAI and its popularity led to addressing negative byproducts of suboptimal and unstructured implementations of prompt patterns a crucial consideration. Hence, the challenges arising due to bad prompt engineering need to be identified and tackled systematically.

\section*{Prompt Smells}

Drawing inspiration from the observed parallels between prompt engineering and coding practices, akin to the concept of code smells, we introduce the notion of prompt smells.

\vspace{0.2cm}
\noindent\fbox{%
\begin{minipage}{.96\columnwidth}
\textbf{Prompt smells are semantic or syntactic characteristics of a prompt instance resulting from (unintentionally) imprecise prompt engineering. Prompt smells can lead to issues related to (i) the desirability of the outputs, (ii) the lack of explainability of the generation process; or (iii) the traces between the input and the output, especially in chain-of-thought strategies.}
\end{minipage}
}\\

Let's take the code for calculating the area of a circle as an example to see different kinds of prompt smells in action. If the prompt is too vague or ambiguous, the generated code may not be what the user intended. For example, if the prompt was ``Write a program that calculates the area of a circle'', the generated code may be in Java instead of Python, affecting the desirability of the output. 

If the prompt is too complicated or convoluted, it would be hard for the user to check which part of the prompt conveyed the user's intentions to the GenAI model. For example, if the prompt was ``Imagine you’re a Python snake slithering around in a circular path. You realize that the path is actually a circle and you’re curious about how much space is inside this path. You remember from your snake school days that the space inside a circle is called the ‘area’. You also remember that there’s a magical number called ‘pi’ that’s roughly 3.14 and that the area can be calculated by multiplying pi with the square of the radius of the circle. Now, as a Python snake, write a Python code that calculates the area of the circle given the radius.'', the generated code might be executable, but the prompt itself is less than ideal as it affects the explainability of the generation process.

If the set of prompts in a chain-of-thought strategy lacks coherence, it would be hard to trace which prompt out of the set of prompts leads to which line of code. For example, if the set of prompts is ``write the area function'', ``write the area parameters'', and ``write the area formula'', this set of prompts is too abstract and vague to trace which prompts leads to which line of code without difficulty.

Table~\ref{tab:cases} outlines the correlation between three key elements: a) the input, which represents the type of dialogue conducted via the prompt (the presence or absence of a prompt smell) used, b) the output, which indicates whether the generated output aligns with the user's expectations (referred to as ``desired'') or deviates from them (referred to as ``undesired''), and c) the case, which classifies each execution into specific scenarios. The cases are divided into three categories: 1) Preferred scenario, where a well-designed dialogue yields the desired output, 2) non-replicable scenarios, where the output is undesired despite utilising a well-designed dialogue or where the output aligns with a user's expectations despite the presence of prompt smells, and 3) standard case scenario where the output is undesired due to the presence of prompt smells. 

Non-replicable cases may occur when the underlying GenAI model exhibits hallucinatory behaviour during specific instances. To detect such hallucinations, a BO3 (best of three) strategy~\cite{ronanki2023chatgpt} can be employed, which involves assessing the semantic consistency of the output across three different and independently conducted dialogue instances. If one instance yields a substantially different output compared to the other two, then we can posit that such an outlier instance might be the result of the GenAI model's hallucination.

\begin{table}[ht!]
    \centering
    \begin{tabular}{|l|l|l|}
        \hline
        \textbf{Input} & \textbf{Output} & \textbf{Case} \\
        \hline
        Well designed & Expected & Preferred scenario\\
        Well designed & Undesired & Non-replicable scenario\\
        Prompt smells & Expected & Non-replicable scenario\\
        Prompt smells & Undesired & Standard case scenario \\
        \hline
    \end{tabular}
    \caption{Cases classified by the quality of the input (prompt instance) and the output of a GenAI}
    \label{tab:cases}
\end{table}

Therefore, the presence of prompt smells, arising from misguided prompt engineering efforts, e.g., poorly implemented prompt patterns or unintentionally imprecise dialogue with a GenAI model, poses a significant challenge to achieving accurate, reliable, and replicable outputs from GenAI models. Identifying prompt smells and using them to trace back the source of the problem will be a major advantage as traceability-enabled explainability and transparency play a crucial role in fostering the trustworthiness of AI systems~\cite{AIact, 10.1145/3597512.3599697}.

\section*{Negative Trends and Future Work}

\emph{Automatic prompting} refers to the process of generating multiple semantically similar prompt candidates for a given task based on output demonstrations. The outputs arising from these instructions are then evaluated to determine the most suitable candidate~\cite{zhou2022large, zhang-etal-2022-promptgen}. We believe that automatic prompting, on top of a lack of prompting standards and transparency mechanisms to identify and mitigate prompt smells, could exacerbate the issue of extrinsic hallucinations in GenAI. As a result, even though automatic prompting can be useful while training GenAI models, it can often nurture prompt smells and increase the emergence of unexpected outcomes. 

It is, therefore, crucial to develop a structured approach to the \emph{art of prompt engineering} and understand the correlations between user intentions, dialogue properties, and GenAI output quality to proactively identify possible prompt smells and mitigate their consequences. Furthermore, the rise of automatic prompting necessitates standardisation despite (or even because of) its innovative nature. By developing a catalogue of prompt smells, we expect to improve the trustworthiness and usability of GenAI models in a wider range of contexts.

\bibliographystyle{ACM-Reference-Format}
\bibliography{main}

\end{document}